\documentclass[11pt,a4paper]{article}
\usepackage[hyperref]{acl2020}
\hypersetup{breaklinks=true}
\usepackage{times}
\usepackage{latexsym}
\usepackage{multirow}
\usepackage{xspace}
\usepackage{mathtools}
\usepackage{bm}
\usepackage{tikz}
\usepackage{pgfplots}
\usepackage{graphics}
\usepackage{todonotes}
\usepackage[ruled,vlined]{algorithm2e}
\renewcommand{\UrlFont}{\ttfamily\small}

\newcommand\BLEU{\textsc{Bleu}\xspace}
\newcommand\TER{\textsc{Ter}\xspace}
\usepackage{microtype}

\aclfinalcopy 

\title{Neural Simultaneous Speech Translation \\ Using Alignment-Based Chunking}

\author{Patrick Wilken, Tamer Alkhouli, Evgeny Matusov, Pavel Golik\\ 
Applications Technology (AppTek), Aachen, Germany \\
{\small \tt \{pwilken,talkhouli,ematusov,pgolik\}@apptek.com } \\
 \\}

\date{}

\begin{document}
\maketitle
\begin{abstract}
In simultaneous machine translation, the objective is to determine when to produce a partial translation given a continuous stream of source words, with a 
trade-off between latency  and quality. We propose a neural machine translation (NMT) model that makes dynamic decisions 
when to continue feeding on input or 
generate output words. The model is composed of two main components: one to dynamically decide on ending a source chunk, and another 
that translates the consumed chunk. We train the components jointly and in a manner consistent with the inference conditions. To generate chunked 
training data, we propose a method that utilizes word alignment while also preserving enough context. We compare models with bidirectional and unidirectional encoders of different depths, both on real speech and text input. 
Our results on the IWSLT\footnote{The International Conference on Spoken Language Translation, \UrlFont http://iwslt.org.} 2020 English-to-German task 
outperform a wait-$k$ baseline by 2.6 to 3.7\% BLEU absolute.
\end{abstract}

\section{Introduction}
Simultaneous machine translation is the task of generating partial translations before observing the entire source sentence. The task fits scenarios such as 
live captioning and  
speech-to-speech translation, where the user expects a translation before the speaker finishes the sentence. Simultaneous 
MT has to balance between latency and translation quality. If more input is consumed before translation, 
quality is likely to improve due to increased context, but latency also increases. On the other hand, consuming limited input decreases latency, but degrades quality. 

There have been several approaches to solve simultaneous machine translation. In \citep{dalvi2018incremental,ma2019stacl}, a fixed policy is 
introduced to delay translation by a fixed number of words.
Alternatively,
\citet{satija2016simultaneous,gu2017learning,alinejad2018prediction} use reinforcement learning to learn a dynamic policy to determine 
whether to read or output words. \citet{cho2016sinmt} adapt the decoding algorithm without relying on additional components.  However, 
these 
methods do not modify the training of the underlying NMT model. 
Instead, it is trained on full sentences. \citet{arivazhagan2019monotonic} introduce a holistic framework that relaxes the hard notion 
of read/write decisions at training time, allowing it to be trained jointly with the rest of the NMT model.

In this paper, we integrate a source chunk boundary detection component into a bidirectional recurrent NMT model. This component corresponds to segmentation or 
read/write decisions in the literature. It is however trained jointly with the rest of the NMT model. We propose an algorithm to chunk the training data based on automatically learned word alignment. 
The chunk boundaries are used as a training signal along with the parallel
corpus. 
The main contributions of this work are as follows:
\begin{itemize}
    \item We introduce a source chunk boundary detection component and train it jointly with the NMT model. Unlike in \citep{arivazhagan2019monotonic}, 
    our component is trained using hard decisions, which is consistent with inference. 
    \item We propose a method based on word alignment to generate the source and target chunk boundaries, which are needed for training.  
    \item We study the use of bidirectional vs unidirectional encoder layers for simultaneous machine translation. Previous work focuses mostly on the use of unidirectional encoders.
    \item We provide results 
    using text and speech input. This is in contrast to previous work that only 
    simulates simultaneous NMT on text input.
\end{itemize}

\section{Related Work}
 \citet{oda2014optimizing} formulate segmentation as an optimization problem solved using dynamic programming to optimize translation quality. 
 The approach is applied to phrase-based machine translation. Our chunking approach is conceptually simpler, and we explore its use with neural machine translation.   
 \citet{cho2016sinmt} devise a greedy decoding algorithm for simultaneous neural machine translation. They use a model that is trained on full sentences. In
 contrast, we train our models on chunked sentences to be consistent with the decoding condition.
 \citet{satija2016simultaneous}, \citet{alinejad2018prediction}, and \citet{gu2017learning} follow a reinforcement learning approach to make decisions 
 as to when to read source words or to write target words.
 \citet{Zheng2019SimplerAF} propose the simpler approach to use the position of the reference target word in the beam of an existing MT system to generate training examples of read/write decisions.
 We extract such decisions from statistical word alignment instead.
 
In \citet{ma2019stacl,dalvi2018incremental}, a wait-$k$ policy is proposed to delay the first target word until $k$ source words are read. The model 
 alternates between generating $s$ target words and reading $s$ source words, until the source words are exhausted. Afterwards, the rest
 of the target words are generated.
In addition, \citet{dalvi2018incremental} convert the training data into  
chunks of predetermined fixed size.  
 In contrast, we train models that learn to produce dynamic context-dependent chunk lengths.

The idea of exploiting word alignments to decide for necessary translation context can be found in several recent papers. \citet{arthur2020learning} train an agent to imitate read/write decisions derived from word alignments. In our architecture such a separate agent model is replaced by a simple additional output of the encoder.
\citet{Xiong2019DuTongChuanCT} use word alignments to tune a pretrained language representation model to perform word sequence chunking. In contrast, our approach integrates alignment-based chunking into the translation model itself, avoiding the overhead of having a separate component and the need for a pretrained model. Moreover, in this work we 
improve on pure alignment-based chunks using language models (Section \ref{subsec:improved_chunking})
to avoid leaving relevant future source words out of the chunk.
\citet{press2018noatt} insert $\epsilon$-tokens into the target using word alignments to develop an NMT model without an attention mechanism. Those tokens fulfill a similar purpose to \textit{wait} decisions in simultaneous MT policies.

\newcite{arivazhagan2019monotonic} propose an attention-based model that integrates an additional monotonic attention component. While the motivation is 
to use hard attention to select the encoder state at the end of the source chunk, they avoid using discrete attention to keep the model differentiable, and use 
soft probabilities instead. The hard mode is only used during decoding. We do not have to work around discrete decisions in this work, since the chunk
boundaries are computed offline before training, resulting in a simpler model architecture. 

\section{Simultaneous Machine Translation}
The problem of offline machine translation is focused on finding the target sequence $e_1^I=e_1...e_I$ of length $I$ given the source sequence $f_1^J$ of 
length $J$. In contrast, simultaneous MT does not necessarily require the full  source input to generate the target output. In this work, 
we formulate the problem by assuming latent monotonic chunking underlying the source and target sequences. 

Formally, let $s_1^K=s_1...s_k...s_K$ denote 
the chunking sequence of $K$ chunks, such that $s_k=(i_k,j_k)$, where $i_k$  denotes the target position of last target word in the $k$-th chunk, and $j_k$ 
denotes the source position of the last source word in the chunk. Since the source and target chunks are monotonic, the beginnings of the source and 
target chunks do not have to be defined explicitly. 
The chunk positions are subject to the following constraints:
\begin{align}
        i_0 = j_0 = 0, & \ \ \ i_K = I, \ \ \ j_K = J, \nonumber \\
        i_{k-1} < i_{k}, & \ \ \  j_{k-1} < j_{k}.
\label{eq:constraints}
\end{align}

We use $\tilde{e}_k=e_{i_{k-1}+1}...e_{i_k}$ to denote the $k$-th target chunk, and 
$\tilde{f}_k=f_{j_{k-1}+1}...f_{j_k}$ to denote its corresponding source chunk. The target sequence $e_1^I$ can be rewritten as $\tilde{e}_1^K$, similarly,
the source sequence can be rewritten as $f_1^J=\tilde{f}_1^K$. 

We introduce the chunk sequence $s_1^K$  as a latent variable as follows:
\begingroup
\allowdisplaybreaks
\begin{align}
&\;p(e_1^I|f_1^J) = \sum_{K,s_1^K} p(e_1^I,s_1^K|f_1^J) \label{eq:latent}\\
               &= \sum_{K,s_1^K} p(\tilde{e}_1^K,s_1^K|\tilde{f}_1^K) \label{eq:rewrite}\\
               &= \sum_{K,s_1^K} \prod_{k=1}^K p(\tilde{e}_k,s_k|\tilde{e}_1^{k-1}, s_1^{k-1}, \tilde{f}_1^K)
               \label{eq:chain} \\
             &= \sum_{K,s_1^K} \prod_{k=1}^K
               p(i_k|\tilde{e}_1^{k}, s_1^{k-1}, j_k, \tilde{f}_1^K) \nonumber \\
             &  \hspace{16.5mm}\cdot p(\tilde{e}_k|\tilde{e}_1^{k-1}, s_1^{k-1}, j_k, \tilde{f}_1^K) \nonumber \\ 
             &  \hspace{16.5mm}\cdot p(j_k|\tilde{e}_1^{k-1}, s_1^{k-1}, \tilde{f}_1^K),   \label{eq:word_chunk_decomp}
\end{align}
\endgroup
where Equation~\ref{eq:latent} introduces the latent sequence $s_1^K$ with a marginalization sum over all possible chunk sequences and all possible number of 
chunks $K$. In Equation~\ref{eq:rewrite} we rewrite the source and target sequences using the chunk notation, and we apply the chain rule of probability 
in Equation~\ref{eq:chain}. We use the chain rule again in Equation~\ref{eq:word_chunk_decomp} to decompose the probability further into a 
\textit{target chunk boundary probability} $p(i_k|\tilde{e}_1^{k}, s_1^{k-1}, j_k, \tilde{f}_1^K)$, a \textit{target chunk translation probability} 
$p(\tilde{e}_k|\tilde{e}_1^{k-1}, s_1^{k-1}, j_k, \tilde{f}_1^K)$, 
and a \textit{source chunk boundary probability} $p(j_k|\tilde{e}_1^{k-1}, s_1^{k-1}, \tilde{f}_1^K)$. This creates a generative story, where
the source chunk boundary is determined first, followed by the translation of the chunk, and finally by the target chunk boundary.  
The translation probability can be further decomposed to reach the word level:
 \begin{align}
 &p(\tilde{e}_k,|\tilde{e}_1^{k-1}, s_1^{k-1}, j_k, \tilde{f}_1^K)  \nonumber \\
     &= \prod_{i=i_{k-1}+1}^{i_k}  p(e_i|\underbrace{e_{i_{k-1}+1}^{i-1}, \tilde{e}_1^{k-1}}_{=e_1^{i-1}}, s_1^{k-1}, j_k,  \tilde{f}_1^K) \nonumber  \\ 
      &\approx  \prod_{i=i_{k-1}+1}^{i_k}  p(e_i|e_1^{i-1}, f_1^{j_k}, k).       
\end{align}

In this work, we drop the marginalization sum over chunk sequences and use fixed chunks during training. The chunk sequences are generated  as described in 
Section \ref{sec:chunking}.
\section{Model}
 \subsection{Source Chunk Boundary Detection}
 \label{sec:boundary_detection}
 We simplify the chunk boundary probability, dropping the dependence on the target sequence and previous target boundary decisions
 \begin{align}
     p(j_k|\tilde{e}_1^{k-1}, s_1^{k-1}, \tilde{f}_1^K)   \approx p(j_k|f_1^{j_k}, j_1^{k-1}),
 \end{align}
 where the distribution is conditioned on the source sequence up to the last word of the $k$-th chunk. It is also conditioned on the previous source 
 boundary decisions $j_1...j_{k-1}$. Instead of computing a distribution over
 the source positions, we introduce a binary random variable $b_j$ such that for each source position we estimate the probability of a chunk boundary:
 \begin{equation}
    b_{j,k} =
    \begin{cases}
        1 &  \text{if\,} j \in \lbrace j_1, j_2 ... j_k \rbrace \\
        0 & \mathrm{otherwise.}
    \end{cases}
 \end{equation}
 
 For this, we use a forward stacked RNN encoder. The $l$-th forward encoder layer is given by
 
 \begin{equation}
    \hspace{-0.05cm}
    \label{eq:forward}
    \overrightarrow{h}_{j,k}^{(l)} = 
    \begin{cases}
      \scalebox{.8}[1.0]{$LSTM$}\big( [\hat{f}_j; \hat{b}_{{j-1},k}], \overrightarrow{h}_{j-1,k}^{(l)}\big) & \hspace*{4mm} l = 1 \vspace{2mm} \\
      \scalebox{.8}[1.0]{$LSTM$}\big( \overrightarrow{h}_{j,k}^{(l-1)}, \overrightarrow{h}_{j-1,k}^{(l)}\big) & \hspace*{-7mm} 1 < l < L_{enc},  
     \end{cases}
 \end{equation}
 
 where $\hat{f}_j$ is the word embedding of the word $f_j$, which is concatenated to the embedding of the  boundary decision at  the previous 
 source position $\hat{b}_{j-1,k}$. $L_{enc}$ is the number of encoder layers.
 On top of the last layer a softmax estimates $p(b_{j,k})$:
 \begin{equation}
    \label{eq:source_chunk_detection}
    \hspace{-0cm}p(b_{j,k}) = \text{softmax}\big(g([\overrightarrow{h}_{j,k}^{(L_{enc})}; \hat{f}_j; \hat{b}_{j-1,k}]) \big),\hspace{-0.2cm}
 \end{equation}
 where $g(\cdot)$ denotes a non-linear function.
 
 \subsection{Translation Model}
 We use an RNN attention model based on~\cite{Bahdanau14:softalign} for $p(e_i|e_1^{i-1}, f_1^{j_k})$. The model shares the forward encoder with the chunk boundary detection model. In addition, we extend the encoder with a stacked backward RNN encoder. The $l$-th backward layer is given by
 \begin{equation}
    \hspace{-0.07cm}
    \label{eq:backward}
    \overleftarrow{h}_{j,k}^{(l)} = 
    \begin{cases}
      \textbf{0} & j>j_k, \forall l \vspace{1.5mm} \\
      \scalebox{.8}[1.0]{$LSTM$}\big( [\hat{f}_j; b_{j,k}], \overleftarrow{h}_{j+1,k}^{(l)}\big) & l = 1  \vspace{2mm} \\  
      \scalebox{.8}[1.0]{$LSTM$}\big( \overleftarrow{h}_{j,k}^{(l-1)}, \overleftarrow{h}_{j+1,k}^{(l)}\big) & \hspace*{-3mm} 1 < l < L_{enc},
     \end{cases}
 \end{equation}
 where the backward layer is computed within a chunk starting at the last position of the chunk $j=j_k$. $\textbf{0}$ indicates a vector of zeros  for positions beyond the current chunk. The source representation is given by the concatenation of the last forward and backward layer
 \begin{equation}
    \label{eq:concat}
     h_{j,k} = [\overrightarrow{h}_{j,k}^{(L_{enc})}; \overleftarrow{h}_{j,k}^{(L_{enc})}].
 \end{equation}
 We also stack $L_{dec}$ LSTM layers in the decoder
 \begin{equation}
    u_{i,k}^{(l)} = 
    \begin{cases}
      \scalebox{.8}[1.0]{$LSTM$}\big( u_{i,k}^{(l-1)}, u_{i-1,\hat{k}}^{(l)}\big) & \hspace*{-2mm} 1 < l < L_{dec} \\
      \scalebox{.8}[1.0]{$LSTM$}\big( [\hat{e}_{i}; d_{i,k}], u_{i-1,\hat{k}}^{(l)}\big) & \hspace*{5mm} l = 1,
     \end{cases}
 \end{equation}
 where $\hat{e}_{i}$ is the target word embedding of the word $e_{i}$, $\hat{k}=k$ unless the previous decoder state belongs to the previous chunk, then 
 $\hat{k}=k-1$. The vector $d_{i,k}$ is the context vector computed over source positions 
 up to the last source position  $j_k$ in the $k$-th chunk
 \begin{align}
     d_{i,k} =& \sum_{j=1}^{j_k} \alpha_{i,j,k} h_{j,k} \\
     \alpha_{i,j,k} =& \text{softmax}(r_{i,1,k}...r_{i,j_k ,k})|_j \\
     r_{i,j,k} =& f(h_{j,k}, u_{i-1,\hat{k}}^{(L_{dec})}),
 \end{align}
 where $\alpha_{i,j,k}$ is the attention weight normalized over the source positions $1\leq j \leq j_k$, and $r_{i,j,k}$ is the energy computed via the 
 function $f$ which uses $\tanh$ of the previous top-most decoder layer and the source representation at position $j$. Note the difference to the 
 attention component  used in offline MT, where the attention weights are computed considering the complete source sentence $f_1^J$.
 The output distribution is computed using a softmax function of energies from the top-most decoder layer $u_{i-1,k}^{(L_{dec})}$, the target embedding
 of the previous word $\hat{e}_{i-1}$, and the context vector $d_{i-1,k}$
 \begin{align}
     p(e_i |& e_1^{i-1}, f_1^{j_k},k) =  \nonumber \\
     &\text{softmax}\big(g([u_{i-1,k}^{(L_{dec})};  \hat{e}_{i-1}; d_{i-1,k}])\big). 
\end{align}

\subsection{Target Chunk Boundary Factor}
Traditionally, the translation model is trained to produce a sentence end token to know when to stop the decoding process. In our approach, this decision has to be made for each chunk (see next section). Hence, we have to train the model to predict the end positions of the chunks on the target side. For this, we use a target factor \citep{garcia2016factored, wilken2019novel}, i.e. a second output of the decoder in each step:
\begin{align}
     \label{eq:target_chunk_detection}
     p(b_i |& e_1^{i}, f_1^{j_k},k) =
     \nonumber \\
     &\text{softmax}(g( u_{i-1,k}^{(L_{dec})}, \hat{e}_{i}, \hat{e}_{i-1}, d_{i-1,k}))
\end{align}
where $b_i$ is a binary random variable representing target chunk boundaries analogous to $b_j$ on the source side. This probability corresponds to the first term in Equation \ref{eq:word_chunk_decomp}, making the same model assumptions as for the translation probability.
Note however, that we make the boundary decision dependent on the embedding $\hat{e}_{i}$ of the target word produced in the current decoding step.

\section{Search}
Decoding in simultaneous MT can be seen as an asynchronous process that takes a stream of source words as input and produces a stream of target words as output. In our approach, we segment the incoming source stream into chunks and output a translation for each chunk individually, however always keeping the full source and target context.

\begin{algorithm}
    \DontPrintSemicolon
    \SetKwInOut{Input}{input}\SetKwInOut{Output}{output}
    \Input{ source word stream $f_1^J$}
    \Output{ target word stream $e_1^I$}
    \BlankLine
    $\hat{\bm{f}_k}$ = [],
    $\overrightarrow{\bm{h}}$ = [],
    $\overleftarrow{\bm{h}}$ = [] \;
    
    \For{$f_j$ \textbf{\upshape in} $f_1^J$}{
        $\hat{f}_j$ = Embedding($f_j$) \;
        $\overrightarrow{h}_j$, $p(b_j)$ = runForwardEncoder($\hat{f}_j$) \;
        $\hat{\bm{f}_k}$ += $\hat{f_j}$ \;
        $\overrightarrow{\bm{h}}$ += $\overrightarrow{h}_j$ \;
        \If{$p(b_i) > t_b$ \textbf{\upshape or} j = J}{
            $\overleftarrow{\bm{h}}_k$ = runBackwardEncoder($\hat{\bm{f}_k}$) \;
            $\overleftarrow{\bm{h}}$ += $\overleftarrow{\bm{h}}_k$ \;
            $\tilde{\bm{e}}_k$ = runDecoder($\overrightarrow{\bm{h}}$, $\overleftarrow{\bm{h}}$) \;
            $e_1^I$ += $\tilde{\bm{e}}_k$ \;
            $\hat{\bm{f}_k}$ = [] \;
            }
 }
 \caption{Simultaneous Decoding \\ \footnotesize{lists in bold, [] is the empty list, += appends to a list}} \label{algo:decoding}
\end{algorithm}

Algorithm \ref{algo:decoding} explains 
the simultaneous decoding process. One source word $f_j$ (i.e. its embedding $\hat{f}_j$) is read at a time. 
We 
calculate the next step of the shared forward encoder (Equation \ref{eq:forward}), including source boundary detection (Equation \ref{eq:source_chunk_detection}). If the boundary probability $p(b_j)$ is below a certain threshold $t_b$, we continue reading the next source word $f_{j+1}$. If, however, a chunk boundary is detected, we first feed all word embeddings $\hat{\bm{f}}_k$ of the current chunk into the backward encoder (Equation \ref{eq:backward}), resulting in representations  $\overleftarrow{\bm{h}}_k$ for each of the words in the current chunk. After that, the decoder is run according to Equations \ref{eq:concat}--\ref{eq:target_chunk_detection}. Note, that it attends to representations $\overrightarrow{\bm{h}}$ and $\overleftarrow{\bm{h}}$ of all source words read so far, not only the current chunk. Here, we perform beam search such that in each decoding step those combinations of target words and target chunk boundary decisions are kept that have the highest joint probability. A hypothesis is considered final as soon as it reaches a position $i$ where a chunk boundary $b_i = 1$ is predicted. Note that 
the length of a chunk translation is not restricted
and hypotheses of different lengths compete. When all hypotheses in the beam are final, the first-best hypothesis is declared as the translation $\tilde{e}_k$ of the current chunk and all its words are flushed into the output stream at once.

During search, the internal states of the forward encoder and the decoder are saved between consecutive different calls while the backward decoder is initialized with a zero state for each chunk.

\section{Alignment-Based Chunking}
\label{sec:chunking}
\subsection{Baseline Approach}
\label{subsec:baseline_chunking}
We aimed at a 
meaningful segmentation of sentence pairs into bilingual chunks which could then be translated in monotonic sequence  
and each chunk is -- in terms of aligned words -- translatable without consuming source words 
from succeeding chunks.
We extract such a segmentation from unsupervised word  alignments in source-to-target and target-to-source directions that we trained using the Eflomal toolkit~\citep{Ostling2016efmaral}
and combined using the \textit{grow-diag-final-and} heuristic~\citep{koehn03:spb}. Then, for each training sentence pair, we extract a sequence of ``minimal-length'' monotonic phrase pairs, i.e. a sequence of the smallest possible bilingual chunks which do not violate the alignment constraints\footnote{This means that all source words within a bilingual chunk are aligned only to the target words within this chunk and vice versa.} and at the same time are 
conform with the segmentation constraints in Equation~\ref{eq:constraints}. By this we allow word reordering between the two languages to happen only within the chunk boundaries. The method roughly follows the approach of~\cite{marino05:tuples}, who extracted similar chunks as units for n-gram based statistical MT.

For fully monotonic word alignments, only chunks of length 1 either on the source or the target side are extracted (corresponding to 1-to-1, 1-to-M, M-to-1 alignments). 
For non-monotonic alignments larger chunks are obtained, in the extreme case the whole sentence pair is one chunk. 
Any unaligned source or target words are attached to the chunk directly preceding them, 
also any non-aligned words that may start the source/target sentence are attached to the first chunk.
We perform the word alignment and chunk boundary extraction on the word level, and then convert words to subword units for the subsequent use in NMT.
\begin{figure*}[t]
\centering
\footnotesize{
\begin{verbatim}
EN:  And | along came | a | brilliant | inventor, | a | scientist, |
     who | came up with a partial cure for that disease
DE:  Dann | kam | ein | brillanter | Erfinder des Wegs, | ein | Wissenschaftler, | 
     der | eine teilweise Heilung für diese Krankheit fand

EN:  And | along came | a brilliant inventor, | a scientist, | 
     who | came up with a partial cure for that disease   
DE:  Dann | kam | ein brillanter Erfinder des Wegs, | ein Wissenschaftler, | 
     der | eine teilweise Heilung für diese Krankheit fand
\end{verbatim}
}
\caption{Examples of the baseline and the improved approach of extracting chunk boundaries. Note how in the improved approach noun phrases were merged into single bigger chunks. Also note the long last chunk that corresponds to the non-monotonic alignment of the English and German subordinate clause.}\label{fig:chunk-example}
\end{figure*}

\subsection{Delayed Source Chunk Boundaries}\label{subsec:delay}

We observed that the accuracy of source boundary detection can be improved significantly by including the words immediately following the source chunk boundary into the context. Take e.\,g. the source word sequence \texttt{I have seen it}. It 
can be translated into German as soon as the word \texttt{it} was read: \texttt{Ich habe es gesehen}. Therefore the model is likely to predict a chunk boundary after \texttt{it}. However, if the next read source word is \textcolor{red}{\texttt{coming}}, it becomes clear that we should have waited 
because the correct German translation is now \texttt{Ich habe es \textcolor{red}{kommen} gesehen}. There is a reordering which invalidates the previous partial translation.

To be able to resolve such cases, we shift the source chunks 
by a constant delay $D$ such that
$j_1, ..., j_k, ..., j_K$ becomes $j_1 + D, ..., j_k + D, ..., j_K + D$.\footnote{If $j_K + D > J$, we shift the boundary to $J$, allowing empty source chunks at sentence end.} Note that the target chunks remain unchanged, thus the extra source words also provide an expanded context for translation. In preliminary experiments we saw large improvements in translation quality when using a delay of 2 or more words, therefore we use it in all further experiments.

\subsection{Improved Chunks for More Context}
\label{subsec:improved_chunking}
The baseline chunking method (Section \ref{subsec:baseline_chunking}) considers word reordering to determine necessary context for translation. However, 
future context is often necessary for correct translation. Consider the translation \texttt{The beautiful woman} $\rightarrow$ \texttt{Die schöne Frau}. Here, despite of the monotonic alignment, we need the context of the third English word \texttt{woman} to translate the first two words as we have to decide on the gender and number of the German article \texttt{Die} and adjective \texttt{schöne}.

In part, this problem is already addressed by adding future source words into the context as described in Section~\ref{subsec:delay}. 
However, this method causes a general increase in latency by $D$ source positions and yet covers only short-range dependencies. A better approach is to remove any chunk boundary for which the words following it are important for a correct translation of the words preceding it. To this end, we introduce a heuristic that uses two bigram target language models (LMs). The first language model yields the probability 
$p(e_{i_k} | e_{i_k - 1})$ for the last word $e_{i_k}$ of chunk $s_k$, whereas the second one computes the probability $p(e_{i_k}|e_{i_k+1})$ for the last word in the chunk given the first word $e_{i_k+1}$ of the next chunk $s_{k+1}$ that follows the word $e_{i_k}$. The chunk boundary after $e_{i_k}$ is removed if the probability of the latter reverse bigram LM is higher than the probability of the first one by a factor $l = \sqrt{i_k - i_{k-1}}$, i.e. dependent on the length of the current chunk. The motivation for this factor is that shorter chunks should be merged with the context to the right more often than chunks which are already long, provided that the right context word has been frequently observed in training to follow the last word of such a chunk candidate. The two bigram LMs are estimated on the target side of the bilingual data, with the second one trained on sentences printed in reverse order.

Examples of the chunks extracted with the baseline and the improved approach for a given training sentence pair are shown in Figure~\ref{fig:chunk-example}.

\section{Streaming Speech Recognition}
\label{sec:asr}
To translate directly from speech signal, we use a cascaded approach. The proposed simultaneous NMT system consumes words from a streaming automatic speech recognition (ASR) system. This system is based on a hybrid LSTM/HMM acoustic model~\citep{bourlard89,Hochreiter:1997:LSTM}, trained on a total of approx.~2300 hours of transcribed English speech from the corpora allowed by IWSLT 2020 evaluation, including MUST-C, TED-LIUM, and LibriSpeech. 
The acoustic model takes 80-dim.~MFCC features as input and estimates state posterior probabilities for 5000 tied triphone states.  It consists of 4 bidirectional layers with 512 LSTM units for each direction. Frame-level alignment and state tying were 
bootstrapped with a Gaussian mixtures acoustic model. 
The LM of the streaming recognizer is a 4-gram count model trained with Kneser-Ney smoothing on English text data (approx.~2.8B running words) allowed by the IWSLT 2020 evaluation. The vocabulary consists of 152K words 
and the out-of-vocabulary rate is below~1\%.
Acoustic training and the HMM decoding were performed with the RWTH ASR toolkit~\cite{wiesler2014:rasr}. 

The streaming recognizer implements a version of chunked processing~\citep{chen2016training,zeyer16} which allows for the same BLSTM-based acoustic model to be used in both offline and online 
applications.
By default, the recognizer updates the current first-best hypothesis by Viterbi decoding starting from the most recent frame and returns the resulting word sequence to the client. This makes 
the first-best hypothesis 
``unstable'', i.e.~past words can change depending on the newly received evidence due to the 
global optimization of the Viterbi decoding. To make the output more stable, we made the decoder delay the recognition results until all active word sequences share a common prefix. 
This prefix is then guaranteed to remain unchanged independent of the rest of the utterance and thus can be sent out to the MT model. 


\section{Experiments}
We conduct experiments on the IWSLT simultaneous translation task for speech translation of TED talks from English to German.
\subsection{Setup}
For training the baseline NMT system, we utilize the parallel data allowed for the IWSLT 2020 evaluation. We divide it into 3 parts: in-domain, clean, and out-of-domain. We consider data from the TED and MUST-C corpora~\citep{mustc19} as in-domain and use it for subsequent fine-tuning experiments, as well as the ``ground truth'' for filtering the out-of-domain data based on sentence embedding similarity with the in-domain data; details are given in \citep{pbahar2020iwsltsystem}. As ``clean'' we consider the News-Commentary, Europarl, and WikiTitles corpora and use their full versions in training. As out-of-domain data, we consider OpenSubtitles, ParaCrawl, CommonCrawl, and \textit{rapid} corpora, which we reduce to 40\% of their total size, or to 23.2M parallel lines, with similarity-based filtering. Thus, in total, we use almost 26M lines of parallel data to train our systems, which amounts to ca.~327M running words on the English side. Furthermore, we added 7.9M sentence pairs or ca.~145M running words of similarity-filtered back-translated\footnote{ The German-to-English system  that we used to translate these data into English is an off-line system trained using the Transformer Base architecture~\citep{transformer} on the in-domain and clean parallel data.} German monolingual data allowed by the IWSLT 2020 evaluation.

In training, the in-domain and clean parallel data had a weight of 5. All models were implemented and trained with the RETURNN toolkit \cite{returnn_acl2018}. We used 
an embedding size of 620 
and LSTM state sizes of 1000.


\begin{table*}[!h]
\centering
\begin{tabular}{llllllll}
\hline
\multirow{2}{*}{\textbf{System}} & \textbf{\hspace*{-2mm} Delay} &\multicolumn{2}{c}{\textbf{tst2015}} & \multicolumn{2}{c}{\textbf{must-c-HE}} & \multicolumn{2}{c}{ \textbf{must-c-COMMON}}    \\
 &&  \textbf{\BLEU} & \textbf{\TER} &  \textbf{\BLEU} & \textbf{\TER} &  \textbf{\BLEU} & \textbf{\TER}   \\
 \hline
\textbf{Offline baseline, Transformer} \\
\hspace{2mm} using reference transcript  & n/a & 32.7 & 50.9 & 30.1 & 54.3 & 32.6 & 48.9 \\
\hspace{2mm} using streaming ASR      & n/a & 28.6 & 56.3 & 26.0 & 59.2 & 26.4 & 57.3 \\
\hline
\textbf{Proposed simultaneous NMT}  & 2 & 24.8 & 60.2 & 21.7 & 63.0 & 21.9 & 60.2 \\
\hspace{2mm} unidirectional,   & 3 & 24.6 & 60.2 & 22.6 & 62.7 & 21.8 & 60.8 \\
\hspace{2mm} (6 enc. 2 dec.)    & 4 & 24.6 & 61.1 & 22.8 & 62.8 & 21.7 & 61.5 \\
\hline
\textbf{Proposed simultaneous NMT} & 2 & 24.6 & 60.0 & 21.4 & 62.8 & 21.9 & 60.6 \\
\hspace{2mm} bidirectional,           & 3 & 24.4 & 60.5 & 22.0 & 62.7 & 21.7 & 61.1 \\
\hspace{2mm} (2x4 enc.\,1 dec.)        & 4 & 24.6 & 61.0 & 21.8 & 63.1 & 21.9 & 61.4 \\
\hline
\end{tabular}
\caption{\label{tbl:speech_mt} Experimental results (in \%) for simultaneous NMT of speech, IWSLT 2020 English$\to$German.}
\end{table*}

As heldout tuning set, we use a combination of IWSLT dev2010, tst2014, and MUST-C-dev corpora. To obtain bilingual chunks as described in Section~\ref{sec:chunking}, we word-align all of the filtered parallel/back-translated and tuning data 
in portions of up to 1M sentence pairs, each of them combined with all of the in-domain and clean parallel data. As heldout evaluation sets, we use IWSLT tst2015, as well as MUST-C HE and COMMON test data. 

For the text input condition, we applied almost no preprocessing, 
tokenization was handled as part of the
subword segmentation with the sentencepiece toolkit~\citep{kudo2018sentencepiece}. The vocabularies for both the source and the target subword models had a size of 30K. For the speech input condition, the additional preprocessing applied to the English side of the parallel data had the goal to make it look like speech transcripts. We lowercased the text, removed all punctuation marks, expanded common abbreviations, especially for measurement units, and converted numbers, dates, and other entities expressed with digits into their spoken form. For the cases of multiple readings of a given number (e.g. \texttt{one oh one}, \texttt{one hundred and one}), we selected one randomly, so that the system could learn to convert alternative readings in English to the same number expressed with digits in German. Because of this preprocessing, our system for the speech condition learned to insert punctuation marks, restore word case, and convert spoken number and entity forms to digits as part of the translation process.

The proposed chunking method (Section \ref{sec:chunking}) is applied to the training corpus as a data preparation step. We measured average chunk lengths of 2.9 source words and 2.7 target words. 40\% of both the source and target chunks consist of a single word, about 20\% are longer than 3 words.

We compute case-sensitive BLEU~\citep{papineni02:bleu} and TER~\citep{snover06:ter} scores as well as the average lagging (AL) metric \citep{ma2019stacl}.

\subsection{Results}
Table \ref{tbl:speech_mt} shows results for the proposed simultaneous MT system. For reference, we first provide the translation quality of an offline system 
that is trained on full sentences. It is a transformer ``base'' model \cite{transformer} that we trained on the same data as the online systems. Row 1 shows BLEU and TER scores for translation of the human reference transcription of the speech input (converted to lower-case, punctuation removed), whereas row 2 uses the automatic transcription generated by our streaming ASR system (Section \ref{sec:asr}). The ASR system has a word error rate (WER) of 8.7 to 11.2\% on the three test sets, causing a drop of 4-6\% BLEU absolute.

All following systems are cascaded streaming ASR + MT online systems that produce translations from audio input in real-time. Those systems have an overall AL of 4.1 to 4.5 seconds, depending on $D$. We compare between two categories of models: unidirectional and bidirectional. For the unidirectional models the backwards decoder (Equation \ref{eq:backward}) was removed from the architecture. We show results for different values of source boundary delay $D$ (see Section \ref{subsec:delay}). For the number of layers we choose $L_\text{enc}=6$ and $L_{\text{dec}}=2$ for the unidirectional models, and $L_{\text{enc}}=4$ (both directions) and $L_\text{dec}=1$ for the bidirectional models, such that the number of parameters is comparable. 
Contradicting our initial assumption, 
bidirectional models do not outperform unidirectional models. This might indicate be due to the fact that the majority of chunks are too short to benefit from a backwards encoding. Also, the model is not sensitive to the delay $D$. This \textit{confirms} our assumption that the additional context of future source words is primarily useful for making the source boundary decision, and for this a context of 2 following (sub-)words is sufficient. For translation, the model does not depend on this ``extra'' context but instead is able to make sufficiently good chunking decisions.

Table \ref{tbl:text_mt} shows results for the case of streamed text input (cased and with punctuation marks). We compare our results to a 4-layer unidirectional system that was trained using the wait-$k$ policy \citep{ma2019stacl}. For this, we chunk the training data into single words, except for a first chunk of size $k=9$ on the source side, and set the delay to $D=0$.
All of our systems 
outperform this wait-$k$ system by large margins. We conclude that the alignment-based chunking proposed in Section \ref{sec:chunking} is able to provide better source context than a fixed policy and that the source boundary detection component described in Section \ref{sec:boundary_detection} successfully learns to reproduce this chunking at inference time. Also for the text condition, we do not observe large differences between uni- and bidirectional models and between different delays.
\begin{table*}[!h]
\centering
\begin{tabular}{lclllllll}
\hline
\multirow{2}{*}{\textbf{System}} & \multirow{2}{*}{\hspace*{-4mm}\textbf{Delay}}& \multirow{1}{*}{\textbf{Avg.}}  & \multicolumn{2}{c}{\textbf{tst2015}} & \multicolumn{2}{c}{\textbf{must-c-HE}} & \multicolumn{2}{c}{\textbf{must-c-COMMON}}    \\
 & & \textbf{AL} &  \textbf{\BLEU} & \textbf{\TER} &  \textbf{\BLEU} & \textbf{\TER} &  \textbf{\BLEU} & \textbf{\TER}   \\
 \hline
\textbf{Baseline} wait-k (k=9)    & - &         & 27.4 & 55.8 & 25.1 & 59.5 & 27.4 & 54.2 \\
\hline
\textbf{Proposed simultaneous MT}                       & 2 & 4.72     & 30.2 & 52.6 & 28.8 & 55.0 & 30.0 & 50.8  \\
\hspace{2mm} unidirectional                     & 3 & 5.26     & 30.3 & 53.2 & 28.8 & 55.2 & 29.7 & 50.8 \\
\hspace{2mm} (6 enc.\,2 dec.)                   & 4 & 6.17     & 29.9 & 53.1 & 28.6 & 55.1 & 29.6 & 50.8 \\
\hline
\textbf{Proposed simultaneous MT}                     & 2 & 4.65     & 29.3 & 53.4 & 28.1 & 55.4 & 29.7 & 50.9 \\
\hspace{2mm} bidirectional                    & 3 & 5.46     & 29.6 & 53.7 & 29.0 & 54.8 & 29.7 & 51.6 \\
\hspace{2mm} (2x4 enc.\,1 dec.)               & 4 & 6.15     & 29.2 & 54.0 & 28.3 & 55.3 & 29.7 & 51.5\\

\hline
\end{tabular}
\caption{\label{tbl:text_mt} Experimental results (in \%) for simultaneous NMT of text input, IWSLT 2020 English$\to$German.}
\end{table*}


\begin{figure}
 \vspace*{-1.5mm}
    \centering
    \begin{tikzpicture}[scale=\textwidth/22cm]
    \begin{axis}[legend pos=north west,ytick distance=1,ymax=31,
        xlabel=AL,
        ylabel=BLEU]
    \addplot[mark=*,blue] plot coordinates {
         (4.6,29.3)
         (5.4,29.6)
         (6.1,29.2)
    };
    \addlegendentry{tst2015}

    \addplot[color=red,mark=x]
        plot coordinates {
             (4.7,28.1)
             (5.6,29.0)
             (6.2,28.3)
        };
    \addlegendentry{mustc-HE}
    
        \addplot[color=black,mark=+]
        plot coordinates {
             (4.64,29.7)
             (5.44,29.7)
             (6.11,29.7)
        };
    \addlegendentry{mustc-COMMON}
    \end{axis}
    \end{tikzpicture}
    \vspace*{-6mm}
    \caption{BLEU vs. AL for bidirectional systems from Table \ref{tbl:text_mt}, generated using a delay $D$ of $2$, $3$, and $4$. }
    \label{fig:bleu_al}
\end{figure}
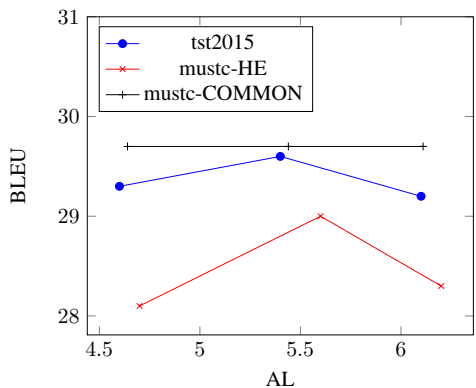

For all systems, we report AL scores averaged over all test sets.
Figure \ref{fig:bleu_al} breaks down the scores to the individual test sets for the bidirectional models.
For a source boundary delay $D=2$ we observe an AL of $4.6$ to $4.7$ words. When increasing $D$, we increase the average lagging score by roughly the same amount, which is expected, since the additional source context for the boundary decision is not translated in the same step where it is added. As discussed before, translation quality does not consistently improve from increasing $D$.

\begin{figure}
    \centering
    \begin{tikzpicture}[scale=\textwidth/22cm]
    \begin{axis}[legend pos=north west,ytick distance=0.5,ymax=29.3,ymin=25.5,
        xlabel=length normalization facor ($\alpha$),
        ylabel=BLEU]
    \addplot[mark=*,blue] plot coordinates {
            (0.5,26.0)
            (0.6,26.9)
            (0.7,27.4)
            (0.8,27.6)
            (0.9,27.8)
            (1.0,27.4)
    };
    \addlegendentry{delay=2}

    \addplot[color=red,mark=x]
        plot coordinates {
            (0.7,28.1)
            (0.8,28.4)
            (0.9,28.5)
            (1.0,28.1)
        };
    \addlegendentry{delay=3}
    
        \addplot[color=black,mark=+]
        plot coordinates {
            (0.7,27.6)
            (0.8,27.9)
            (0.9,28.2)
            (1.0,28.0)
        };
    \addlegendentry{delay=4}
    \end{axis}
    \end{tikzpicture}
    \vspace*{-6mm}
    \caption{BLEU vs. the length normalization factor ($\alpha$) on the tuning set (dev2010 + tst2014 + MUST-C-dev).}
    \label{fig:len_norm_tuning}
\vspace*{-5mm}    
\end{figure}
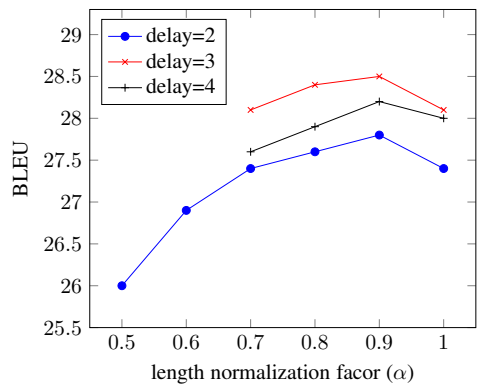

We found tuning of length normalization to be important, as the average decoding length for chunks is much shorter than in offline translation. For optimal results, we divided the model scores by $I^\alpha$, $I$ being the target length, and tuned the parameter $\alpha$. Figure \ref{fig:len_norm_tuning} shows that  $\alpha=0.9$ works best in our experiments, independent of the source boundary delay $D$. This value is used in all experiments.

Furthermore, we found the model to be very sensitive to a source boundary probability threshold $t_b$ different than $0.5$ regarding translation quality. This means the ``translating'' part of the network strongly adapts to the chunking component.

\section{Conclusion}
We proposed a novel neural model architecture for simultaneous MT that incorporates a component for splitting the incoming source stream into translatable chunks. We presented how we generate training examples for such chunks from statistical word alignment and how those can be improved via language models. Experiments on the IWSLT 2020 English-to-German task proved that the proposed learned source chunking outperforms a fixed wait-$k$ strategy by a large margin. We also investigated the value of backwards source encoding in the context of simultaneous MT by comparing uni- and bidirectional versions of our architecture.
\section*{Acknowledgements}
We would like to thank our colleague Albert Zeyer for fruitful discussions and RETURNN implementation support.

\bibliography{translation,anthology}
\bibliographystyle{acl_natbib}

\vfill
\end{document}